\pdfoutput=1

\documentclass[11pt]{article}

\usepackage[]{acl}

\usepackage{times}
\usepackage{latexsym}

\usepackage[T1]{fontenc}

\usepackage[utf8]{inputenc}

\usepackage{microtype}
\usepackage{comment}

\usepackage[hide]{boxnotes} 
\usepackage{soul} 

\usepackage{booktabs}
\usepackage{xcolor}
\definecolor{ggreen}{RGB}{26,168,65}
\definecolor{bblue}{RGB}{11,125,212}

\title{Automatic Detection of Entity-Manipulated Text Using Factual Knowledge}

\author{Ganesh Jawahar$^{\dag,\ddag}$  ~~~Muhammad Abdul-Mageed$^\dag$
~~~ Laks V. S. Lakshmanan$^{\dag,\ddag}$\\
\normalsize $^\dag$Deep Learning \& Natural Language Processing Group, $^\ddag$Data Management \& Mining Group  \\
\normalsize The University of British Columbia\\
  \texttt{ \small ganeshjwhr@gmail.com,\{laks,amuham01\}@cs.ubc.ca} }

\begin{document}
\maketitle

\begin{abstract}
In this work, we focus on the problem of distinguishing a human written news article from a news article that is created by manipulating entities in a human written news article (e.g., replacing entities with factually incorrect entities). Such manipulated articles can mislead the reader by posing as a human written news article. 
We propose a neural network based detector that detects manipulated news articles by reasoning about the facts mentioned in the article. Our proposed detector exploits factual knowledge via graph convolutional neural network 
along with the textual information in the news article. 
We also create challenging datasets for this task by considering various strategies to generate the new replacement entity (e.g., entity generation from GPT-2). In all the settings, our proposed model either matches or outperforms the state-of-the-art detector in terms of accuracy. Our code and data are available at \url{https://github.com/UBC-NLP/manipulated_entity_detection}. 
\end{abstract}
\section{Introduction}
\label{sec:intro} 
A type of fake news that has received little attention in the research community is manipulated text. Manipulated text is typically created by manipulating a human written news article minimally (e.g., replacing every occurrence of a particular entity, `Obama' in a news article with another American politician entity). Current fake news detectors that exploit stylometric signals from the text (e.g., choice of specific words to express false statements) are clearly insufficient for distinguishing manipulated text from human written text~\cite{ZhouGBH19,schuster_cl20} as the style underlying the manipulated text is virtually identical to human writing style. 
In this work, we focus on this problem of distinguishing manipulated news articles from human written news articles.

We consider a particular type of text manipulation --- entity perturbation~\cite{ZhouGBH19}, where a manipulated news article is created by modifying a fixed number of entities in a human written news article (e.g., replacing them with entities generated from a  text generative model). 
\begin{table}
\centering
\small
\begin{tabular}{p{2.8in}} \toprule
\textbf{Human written text} \\
\textbf{\textcolor{ggreen}{PubNub}}, a startup that develops the infrastructure to power key features in real-time applications (...) has raised \$23 million in a series D round of funding from \textbf{\textcolor{ggreen}{Hewlett Packard Enterprise (HPE)}}, \textbf{\textcolor{ggreen}{Relay Ventures}}, \textbf{\textcolor{ggreen}{Sapphire Ventures}}, \textbf{\textcolor{ggreen}{Scale Venture Partners}}, \textbf{\textcolor{ggreen}{Cisco Investments}}, \textbf{\textcolor{ggreen}{Bosch}}, and \textbf{\textcolor{ggreen}{Ericsson}}. \\ \midrule 
\textbf{Manipulated text using GPT-2} \\
\textbf{\textcolor{ggreen}{PubNub}}, a startup that develops the infrastructure to power key features in real-time applications (...) has raised \$23 million in a series D round of funding from \textbf{\textcolor{ggreen}{Hewlett Packard Enterprise (HPE)}}, \textbf{\textcolor{orange}{Samsung}}, \textbf{\textcolor{ggreen}{Sapphire Ventures}}, \textbf{\textcolor{ggreen}{Scale Venture Partners}}, \textbf{\textcolor{ggreen}{Cisco Investments}}, \textbf{\textcolor{ggreen}{Bosch}}, and \textbf{\textcolor{ggreen}{Ericsson}}.\\ \bottomrule
\end{tabular}
\caption{Example human written and manipulated text. Named entities of organization type are shown in \textbf{\textcolor{ggreen}{green}}. Manipulated entities are shown in \textbf{\textcolor{orange}{orange}}.}
\label{tabl:sample_human_text}
\end{table}
E.g., in Table~\ref{tabl:sample_human_text}, to mislead humans, the entity `Relay Ventures' can be replaced by `Samsung' (a candidate replacement entity generated by the generative pre-training-2 model (GPT-2)~\cite{radford2019language}), which is locally consistent as some of the other companies in the original text are also into device manufacturing.

To distinguish a manipulated news article from the original human written news article, we propose a neural network based detector that jointly utilizes the textual information along with the the factual knowledge explicitly by building entity-relation graphs which capture the relationship between different entities present in the news article. The factual knowledge is encoded by a graph convolutional neural network~\cite{kipf2017semi} that captures the interactions between different entities and relations, which we hypothesize, carries discriminatory signals for the manipulated text detection task. 
Our major contributions include: (i) a detector that exploits factual knowledge to overcome the limitations of relying only on stylometric signals, 
(ii) an approach to generate challenging manipulated news article dataset using GPT-2, and (iii) a collection of challenging datasets by considering various strategies to generate the replacement entity.

\section{Background and Related Work}
\label{sec:relatedwork}
The manipulated text detection task is related to diverse research areas such as fake news detection, natural language understanding, and knowledge bases.

\noindent\textbf{Fake news detection.}
Research on Fake news detection typically deals with challenges such as understanding the news content~\cite{schuster_cl20}, claim verification~\cite{thorne-vlachos-2018-automated}, verifying the credibility of the source~\cite{castillo_infocred}, and exploiting fake news propagation  patterns~\cite{vosoughi2018spread}. Our work is primarily focused on detecting fake news in the form of manipulated text, by understanding the news content.
In the traditional problem setting, both fake and real news is assumed to be written by a  human~\cite{fn_survey_dmp,oshikawa-etal-2020-survey}. 
Since humans tend to make stylistic choices (e.g., choosing some specific language for writing false statements), the fake news detector can perform reasonably on the task by picking up on these stylometric signals. One can also create fake news by manipulating a human written news article minimally. Such manipulations include: entity perturbation (e.g., `12 people were injured in the shooting' to `24 people were killed in the shooting')~\cite{ZhouGBH19}, 
subject-object exchange (e.g., `A gangster was shot by the police' to `A policeman was shot by the gangster')~\cite{ZhouGBH19}, and
adding/deleting negations (e.g., `Trump doesn't like Obamacare' to `Trump likes Obamacare')~\cite{schuster_cl20}. These manipulations do not typically affect the style and hence stylometric signals alone cannot help in building accurate manipulated text detection models~\cite{ZhouGBH19,schuster_cl20}. 

\noindent\textbf{Natural language understanding.}
Pre-trained language models such as BERT~\cite{devlin_naacl19} and  RoBERTa~\cite{liu_roberta19} achieve strong performance in diverse NLP tasks. 
Specifically, RoBERTa 
is the state-of-the-art detector when finetuned for detection of synthetic text~\cite{solaiman_arxiv19,jawahar-etal-2020-automatic}. These models can also capture implicit world knowledge (e.g., Paris is the capital of France) 
that occurs frequently in the  text~\cite{petroni-etal-2019-language}. 
However, it is insufficient for solving our task~\cite{schuster_cl20}, as it is limited to frequent patterns.

\noindent\textbf{Knowledge bases (KBs).} 
Knowledge bases (e.g., YAGO~\cite{YAGO4}) containing typically a collection of facts (e.g., subject-relation-object triples), provide specialized knowledge for downstream NLP tasks (e.g., question answering~\cite{banerjee-baral-2020-self}). 
One can integrate such symbolic knowledge into pre-trained language models during pre-training~\cite{zhang-etal-2019-ernie} and finetuning (\newcite{weijie2019kbert,zhong-etal-2020-neural}, which we follow in this work). 
\section{Manipulated Text Creation}
\label{sec:man_text_gen}

In this work, we focus on a particular type of manipulation --- entity perturbation~\cite{ZhouGBH19}, where all occurrences of a fixed number of randomly picked entities from a human written news article are replaced with different replacement entities. We replace named entities of three types: person, organization and location (recognized using spaCy's named entity recognizer (NER)~\cite{spacy}). We ensure the replacement (new) entity belongs to the same type as the original (old) entity. We create challenging manipulated text datasets by considering various strategies to identify the new replacement entity: random most frequent entity  (pick randomly from among the top 5000 entities), random least frequent entity  (pick randomly from the bottom 5000 entities), and entity generated by GPT-2. Sample manipulated entities obtained from different replacement strategies are shown in Table~\ref{tab:sample_entities}. 

\begin{table}[htb]
\scriptsize 
\begin{center}
\begin{tabular}{p{0.85in}p{0.8in}p{0.89in}} 
\toprule
\multicolumn{3}{c}{\textbf{Entity replacement strategy}} \\ \midrule 
\textbf{Random least} & \textbf{Random most} & \textbf{GPT-2 generated}  \\ \midrule
Inverkeithing High School & Tribune & U.S. \\
Mark Forman & East Jerusalem & Canada \\
Netgear & Englishman & Microsoft \\
Bangalore North & Jason Aldean & Donald Trump \\
Mackintosh & UFA & BBC  \\ 
\bottomrule
\end{tabular}
\caption{Sample manipulated entities}
\label{tab:sample_entities}
\end{center}
\end{table}

\noindent\textbf{GPT-2 generated entity replacement.}
Strategies that randomly identify the replacement entity ignore the context provided by the news article. For example, in news portion (\ref{tabl:sample_human_text}), a random replacement entity for `Relay Ventures' can be `Salesforce'. However, it is likely locally inconsistent as `Salesforce' is not into device manufacturing unlike many other co-occurring companies in the original text. We propose a novel approach that makes use of the state-of-the-art text generative model GPT-2 to pick replacement entities that are locally consistent. Revisiting the news portion (\ref{tabl:sample_human_text}), let the randomly selected entity to be replaced be `Relay Ventures'.  \note[Laks]{Can GPT-2 be called SOTA, given GPT-3?}
We treat the fragment of text from the beginning of the article up to the tokens before the first occurrence of the target entity (`Relay Ventures') as the prompt. We provide this prompt to GPT-2, which can then generate the next few tokens.
We call the generated token sequence a candidate replacement entity if the sequence starts with an entity (e.g., `Samsung') of same type  as the target entity (`Relay Ventures') and has no string overlap with the target entity.
If the constraints are not met, we ask GPT-2 to create the generated sequence again up to a maximum of 10 attempts. The candidate replacement entity thus obtained will be used to replace all occurrences of the target entity. For the news portion (\ref{tabl:sample_human_text}), the candidate replacement entity generated by GPT-2 is `Samsung', which is locally consistent: similar to other companies in the original text, Samsung manufactures devices.
\section{Manipulated Text Detection}
\label{sec:mtd}
The goal of this work is to build a detector that distinguishes manipulated news article from human written news article with high accuracy. In prior work, \newcite{ZhouGBH19} conclude that the manipulated article can possibly be detected by checking the facts underlying the article with knowledge bases and  \newcite{schuster_cl20} show that humans can identify the manipulated text well when they are allowed to consult external sources (e.g., internet). Building on these findings, we hypothesize that \textit{factual knowledge underlying the news article can provide discriminatory signals for manipulated text detection.} To this end, we embody the RoBERTa detector with explicit factual knowledge so that the detector can reason about facts present in the news article, whose details we discuss next.

\begin{table*}[htb]
\scriptsize
\begin{center}
\begin{tabular}{p{1.8in}p{0.19in}p{0.19in}p{0.19in}p{0.21in}p{0.21in}p{0.21in}p{0.20in}p{0.20in}p{0.20in}} \toprule 
\textbf{Entity replacement strategy} & \multicolumn{3}{p{1.2in}}{\textbf{Random least frequent entity replacement}} & \multicolumn{3}{p{1.25in}}{\textbf{Random most frequent entity replacement}} & \multicolumn{3}{p{1.15in}}{\textbf{GPT-2 generated entity replacement}} \\ \midrule 
Maximum no. of entity replacements & 1 & 2 & 3 & 1 & 2 & 3 & 1 & 2 & 3 \\ \midrule 
\textbf{Manipulated Article Detection Task} \\
\textit{(1) Overall Accuracy} \\
RoBERTa & 67.09 & 78.37 & 84.26 & 65.56 & 76.86 & 83.93 & 67.09 & 74.12 & 78.79 \\
Ours (w/o Entity Identification Objective) & 67.25 & 78.36 & \textbf{84.59} & $66.99^{*}$  & $77.98^{*}$  & 83.86 & \textbf{67.16} & 73.84 & \textbf{79.11} \\
Ours & $\textbf{68.25}^{*}$ & \textbf{78.99} & 83.84 & $\textbf{67.21}^{*}$ & $\textbf{78.26}^{*}$ & \textbf{84.39} & 65.84 & \textbf{74.80} & \textbf{79.05}  \\ \hline
\textbf{Manipulated Entity Identification Task} \\
\textit{(1) Overall Precision} - Ours & 49.99 & 50.02 & 50.08 & 49.94 & 50.00 & 49.83 & 49.49 & 48.52 & 48.71 \\ 
\textit{(2) Overall Recall} -
Ours & 38.56 & 55.11 & 65.11 & 48.20 & 50.04 & 47.71 & 45.82 & 46.76 & 45.67 \\ 
\textit{(3) Overall F-Score} -
Ours & 42.29 & 46.50 & 46.12 & 46.07 & 47.79 & 46.83 & 44.82 & 47.42 & 44.92 \\ 
\textit{(4) Manipulated Entity - Precision} -
Ours & 81.06 & 91.76 & 84.14 & 84.71 & 88.06 & 86.06 & 85.59 & 85.91 & 73.80 \\ 
\textit{(5) Manipulated Entity - Recall} -
Ours & 0.00 & 3.70 & 12.12 & 6.08 & 4.63 & 14.03 & 9.14 & 1.64 & 12.50 \\ 
\textit{(6) Manipulated Entity - F-Score} -
Ours & 0.00 & 7.11 & 21.19 & 11.35 & 8.80 & 24.13 & 16.52 & 3.22 & 21.38 \\ \bottomrule 
\end{tabular}
\caption{Evaluation performance (\%) for different maximum number of entity replacements across different replacement strategies. \textbf{Bolded} refers to the best results for each dataset. Note that the state-of-the-art detector cannot identify manipulated entities present in the document. For the manipulated article detection task, statistically significant overall accuracy results obtained using bootstrap test with $p < 0.01$ are marked using asterisk ($^{*}$).}
\label{tab:fullres}
\end{center}
\end{table*}

\noindent\textbf{Factual knowledge.} For factual knowledge, we leverage a variant of YAGO 4 KB~\cite{YAGO4} that contains only instances that have an English Wikipedia article. We then extract the facts in a given document by first identifying all the entities present in the document using spaCy's NER. For each target entity, we grab all the triples in the KB where the subject matches with the target entity at surface level. These triples can  be seen as the first hop neighbors of the target entity in the KB. For a given document, the set of triples collected over all identified entities is used to build the corresponding factual graph. A node can be an entity or a relation. A directed edge is added between subject and relation, as well as relation and object. This factual graph  contains rich factual information about entities present in the document, which can be exploited to reason about facts mentioned in the article for correctness. 
\noindent\textbf{Integrating factual knowledge with RoBERTa.}
Our proposed detector is an integration of the RoBERTa model with factual knowledge. This allows the detector to reason about facts mentioned in the article. To embed the factual knowledge, we employ graph convolutional networks (GCNs)~\cite{kipf2017semi}, where we stack $l$ GCN layers and the definition of the hidden representation of each node $v$ of the factual graph as layer $k+1$, in a graph $\mathcal{G}=(\mathcal{V},\mathcal{E})$:
\begin{equation}\label{eq:gcn-khop} 
\small
\mathbf{h}_v^{k+1} = f\left(\frac{1}{|\mathcal{N}(v)|} \sum_{u \in \mathcal{N}(v)} \mathcal{W}^kh_u^k + b^k\right), \quad  \forall v \in \mathcal{V},
\end{equation}
\noindent where $\mathcal{W}^k$, $b^k$, $h_u^k$, $\mathcal{N}(v)$ correspond to layer specific model weights, biases, node representation, and neighbors of v in $\mathcal{G}$ respectively. Note that $h_u^1$ denotes the initial node features, which can be initialized randomly or using a pre-trained entity embedding such as Wikipedia2vec~\cite{wikipedia2vec}.

\noindent\textbf{Detector prediction.} 
The factual knowledge about entities present in the article is captured in the node embeddings ($h_u^l$) corresponding to the last layer $l$ of the GCN model. The textual knowledge corresponding to the document can be obtained from the last layer representation ($r_{CLS}^d$) of the RoBERTa model corresponding to the first token (`[CLS]', special classification token) of the RoBERTa input. 
We combine the factual and the textual knowledge by simply averaging all the GCN's entity embeddings and concatenating the entity average with the RoBERTa's document embedding. Thus, the unnormalized prediction probabilities ($mf(d)$) of our detector for the document $d$ can be given by:
\begin{equation}\label{eq:dpred} 
\small
\mathbf{mf}(d) = \mathcal{W}_{mtd} \left[ r_{[CLS]}^d ; \sum_{e \in entities(d) } h^l_e \right]   + b_{mtd},
\end{equation}
\noindent where $[;]$ corresponds to the concatenation operation and $\mathcal{W}_{mtd}$, $b_{mtd}$ correspond to the affine transformation specific model parameters for manipulated text detection. The output from $\mathit{mf}(d)$ passes through dropout followed by ReLU layer. 

\noindent\textbf{Identifying manipulated entities.} 
To enable humans to understand our detector's decision and perform further investigation, we introduce a subtask for the detector, namely identify the manipulated entities among different entities present in the document. For this subtask, we build on the entity representations output by the last layer of the GCN model. The unnormalized class prediction  probabilities ($ef(v)$) for a given entity $v$ from the article can be given by:
\begin{equation}\label{eq:manent} 
\small
\mathbf{ef}(v) = Dropout\left(ReLU\left(\mathcal{W}_{ec} h_v^l + b_{ec}\right)\right),
\end{equation}
\noindent where $h_v^l$ denotes the hidden representation at last layer $l$ for the entity $v$, and $\mathcal{W}_{ec}$, $b_{ec}$ correspond to the affine transformation specific model parameters for entity classification. 
The overall objective function of the proposed detector can be given by:
\begin{equation}\label{eq:ovfn} 
\scriptsize
\min_{\theta}  \sum_{i=1}^n \left[ \mathcal{L}(s(mf(x_i)), y_i) + \sum_{e\in entities(x_i)} \mathcal{L}(s(ef(e)), y^e) \right].
\end{equation}
\noindent where $\mathcal{L}$, $mf$, and $s$ resp. correspond to the function that computes the negative log-probability of the correct label, detection prediction function, and softmax function. $y^e$ denotes the entity manipulation class label, which is $1$ if the entity $e$ is manipulated, and $0$ otherwise. $y_i$ denotes the article manipulation class label, which is $1$ if at least one entity in article $i$ is manipulated, and $0$ otherwise.

\section{Experiments and Results}
\label{sec:expres}

\begin{table*}[htb]
\scriptsize  
\begin{center}
\begin{tabular}{p{1.75in}p{0.20in}p{0.20in}p{0.20in}p{0.20in}p{0.20in}p{0.20in}p{0.20in}p{0.20in}p{0.20in}} \toprule 
\textbf{Entity replacement strategy} & \multicolumn{3}{p{1.25in}}{\textbf{Random least frequent entity replacement}} & \multicolumn{3}{p{1.15in}}{\textbf{Random most frequent entity replacement}} & \multicolumn{3}{p{1.15in}}{\textbf{GPT-2 generated entity replacement}} \\ \midrule 
Maximum no. of entity replacements & 1 & 2 & 3 & 1 & 2 & 3 & 1 & 2 & 3 \\ \midrule 
Test set size (Percent) & 3,797 (47.5) & 3,625 (45.3) & 3,447 (43.1) & 3,288 (41.1) & 2,660 (33.2) & 2,207 (27.6) & 3,302 (41.3) & 2,737 (34.2) & 2,359 (29.5) \\ \midrule
RoBERTa & 48.17 & 68.69 & 77.81 & 45.62 & 66.32 & 74.94 & 51.97 & \textbf{66.97} & \textbf{74.95} \\
Ours (w/o Entity Identification Objective) & 47.20 & 65.19 & \textbf{78.76} & 51.55 & 68.20 & \textbf{75.44} & 56.27 & 66.68 & 72.11  \\
Ours & \textbf{52.04} & \textbf{68.99} & 75.98 & \textbf{54.65} & \textbf{68.38} & 72.81 & \textbf{62.11} & 66.53 & 71.22  \\ \bottomrule 
\end{tabular}
\caption{Manipulated article detection performance (\%) for different maximum number of entity replacements across different replacement strategies on a subset of our test set. This text subset contains manipulated articles with all the manipulated entities absent in the knowledge base. \textbf{Bolded} refers to best results for each dataset.}
\label{tab:entity_ood}
\end{center}
\end{table*}


\noindent\textbf{Dataset and Detector Settings.} 
The human written news articles used in our study are  taken from the RealNews dataset~\cite{zellers_neurips19}, which contains $5000$, $2000$, and $8000$ news articles in the training, validation, and test set respectively. We randomly pick half of the news articles in each set for human written news article category and the rest in each set for manipulation based on the chosen replacement strategy. We also create three different datasets for each replacement strategy by varying the maximum number of entities to be manipulated from 1 to 3. 
Detailed statistics of the proposed datasets is in \ref{sec:stats}. The hyperparameter search space for all detectors is offered in \ref{sec:hyp_space}.

\noindent\textbf{Hardest detection task.} Table~\ref{tab:fullres} presents the detection accuracy results. We observe that the most challenging dataset for the state-of-the-art detector is surprisingly from random most frequent entity replacement strategy with exactly one entity replacement. The random strategies fail to create a challenging dataset with high (e.g., 3) number of entity replacements, which indicates that the detection task becomes easier with increase in the number of locally inconsistent entities. Nevertheless, our proposed GPT-2 based entity replacement strategy keeps the detection task harder even for large number of replacements, thanks to the ability of the strategy to generate locally consistent entities mostly. Regardless of the replacement strategies, the detection performance of all the detectors increases with the increase in the number of entities that are manipulated in a document, that is, more the manipulations in a document, the easier the detection task. This result is similar to  previous research which performs manipulation by adding/deleting negations in  news articles~\cite{schuster_cl20}. A fake news propagator can thus manipulate exactly one entity in the news article to make the detection task harder.

\noindent\textbf{Detector performance.} Nevertheless, our proposed detector performs similarly to or outperforms the state-of-the-art detector on all replacement strategies across different numbers of entity replacements. This result validates our hypothesis that leveraging both factual and textual knowledge can improve detection performance, overcoming the limitations of relying only on textual knowledge. Improvements of our proposed detector on the GPT-2 generated entity manipulation task are not significantly high due to sizeable increase in manipulated entities absent in the knowledge base (e.g., $\sim$50\%, see last three rows in Table~\ref{tab:dataset_stats}).

\noindent\textbf{Entity identification performance.} Our proposed detector is equipped to identify entities that are manipulated in a news article. This task is harder due to the imbalanced nature of the task as most of the entities present in the news article are not manipulated. As shown in Table~\ref{tab:fullres}, our proposed detector achieves high precision ($\geq 70\%$) in identifying manipulated entities, which makes our detector appealing for applications that favor precision. The recall is very low ($<15\%$), which indicates the difficulty of the task. We also experiment with a baseline RoBERTa model trained at the token level to identify spans of manipulated entities. However, the model seems overwhelmed by the majority class (token not part of the manipulated entity span) and predicts all the tokens to belong to the majority class. We believe there is a lot of room for improvement in this subtask. 

\noindent\textbf{Detecting articles with unknown manipulated entities.} Table~\ref{tab:entity_ood} shows performance of the detector on manipulated articles when all the manipulated entities are not present in the knowledge base. We observe that our proposed detector can rely on the relations corresponding to the non-manipulated entities and pretrained textual representations to outperform, or at least be on par with, the RoBERTa model.

\begin{table}[htb]
\scriptsize
\begin{center}
\begin{tabular}{cccc} 
\toprule
\textbf{Repl. strategy} / \# replacements & 1 & 2 & 3 \\ \midrule 
Random least frequent & 93.67 & 95.06 & 95.05 \\
Random most frequent & 93.75 & 93.37 & 93.79 \\
GPT-2 generated & 95.1 & 93.35 & 94.88 \\ \bottomrule
\end{tabular}
\caption{Quality gap - Human vs. Manipulated text}
\label{tab:muave_results}
\end{center}
\end{table}

\noindent\textbf{Quality gap between human and manipulated text.} Table~\ref{tab:muave_results} shows how the quality of the manipulated text changes with respect to human written text across different replacement strategies, for different numbers of replacements. We utilize MAUVE~\cite{pillutla2021mauve}, a metric to measure the closeness of machine generated text to human language based on divergence frontiers. Since the proposed manipulations touch only limited spans (i.e., entities) in the entire document, the overall quality of the manipulated text does not change much with more replacements.

\section{Conclusion}
\label{sec:conclusion}
We presented the first principled approach for developing a model that can detect entity-manipulated text articles.  
In addition to textual information, our proposed detector exploits explicit factual knowledge from a knowledge base to overcome the limitations of relying only on stylometric signals. We constructed challenging manipulated datasets by considering various entity replacement strategies, including with random selection and GPT-2 generation. On all the experimental settings, our proposed model outperforms (or is at least on par with) the baseline detector in overall detection accuracy. 
Our results show that manipulated text detection remains challenging. 
We hope that our work will trigger further research on this important but relatively understudied subfield of fake news detection. 

\section*{Acknowledgements}\label{sec:acknow}
We gratefully acknowledge support from the Natural Sciences and Engineering Research Council of Canada (NSERC; RGPIN-2018-04267), the Social Sciences and Humanities Research Council of Canada (SSHRC; 435-2018-0576), Canadian Foundation for Innovation (CFI; 37771), Compute Canada (CC),\footnote{\url{https://www.computecanada.ca}{https://www.computecanada.ca}} UBC ARC-Sockeye,\footnote{\url{https://arc.ubc.ca/ubc-arc-sockeye}{https://arc.ubc.ca/ubc-arc-sockeye}} and Advanced Micro Devices, Inc. (AMD). Any opinions, conclusions or recommendations expressed in this material are those of the author(s) and do not necessarily reflect the views of NSERC, SSHRC, CFI, CC, ARC-Sockeye, or AMD. We also thank Ayushi Dalmia for proofreading and helpful discussions.

\bibliography{anthology,custom}
\bibliographystyle{acl_natbib}

\pagebreak

\appendix
\section{Appendices}
\label{sec:appendix}


\subsection{Summary Statistics of Proposed Datasets.}
\label{sec:stats}
Table~\ref{tab:dataset_stats} displays the statistics of proposed datasets.

\begin{table*}[htb]
\footnotesize
\begin{center}
\begin{tabular}{p{2.0in}p{0.20in}p{0.20in}p{0.20in}p{0.20in}p{0.20in}p{0.20in}p{0.20in}p{0.20in}p{0.20in}} \toprule 
\textbf{Name} & \multicolumn{3}{p{1.25in}}{\textbf{Random least frequent entity replacement}} & \multicolumn{3}{p{1.15in}}{\textbf{Random most frequent entity replacement}} & \multicolumn{3}{p{1.15in}}{\textbf{GPT-2 generated entity replacement}} \\ \midrule 
 Maximum no. of entity replacements & 1 & 2 & 3 & 1 & 2 & 3 & 1 & 2 & 3 \\ \midrule 
\textit{Dataset Size} \\
Train & 5,000 & 5,000 & 5,000 & 5,000 & 5,000 & 5,000 & 5,000 & 5,000 & 5,000  \\
Validation & 2,000 & 2,000 & 2,000 & 2,000 & 2,000 & 2,000 & 2,000 & 2,000 & 2,000  \\
Test & 8,000 & 8,000 & 8,000 & 8,000 & 8,000 & 8,000 & 8,000 & 8,000 & 8,000  \\  
\textit{Average Length (\# words)} \\
Train  & 604 & 604 & 605 & 603 & 603 & 603 & 603 & 613 & 614  \\
Validation & 595 & 595 & 596 & 594 & 594 & 594 & 607 & 598 & 599 \\
Test & 597 & 597 & 597 & 596 & 596 & 596 & 598 & 598 & 601 \\
\textit{\% Documents with Person Entities  } \\
Train  & 97.92 & 98.00 & 97.96 & 97.74 & 97.84 & 98.00 & 97.22 & 97.60 & 97.82   \\
Validation & 98.65 & 98.65 & 98.85 & 98.55 & 98.65 & 98.50 & 97.80 & 98.00 & 98.30 \\
Test & 97.86 & 98.04 & 98.16 & 97.92 & 97.91 & 97.95 & 97.45 & 97.49 & 97.76 \\
\textit{\% Documents with Organization Entities} \\
Train  & 99.14 & 99.12 & 99.10 & 99.20 & 99.26 & 99.12 & 99.04 & 99.10 & 99.14   \\
Validation & 99.35 & 99.35 & 99.30 & 99.35 & 99.35 & 99.40 & 99.20 & 99.50 & 99.25 \\
Test & 99.28 & 99.20 & 99.17 & 99.24 & 99.12 & 99.17 & 99.06 & 99.05 & 99.11 \\
\textit{\% Documents with Location Entities} \\
Train  & 90.44 & 90.16 & 89.84 & 90.70 & 90.70 & 91.00 & 90.70 & 91.34 & 91.88   \\
Validation & 90.40 & 89.90 & 89.75 & 90.55 & 90.55 & 90.80 & 90.80 & 91.05 & 91.90 \\
Test & 90.69 & 90.28 & 89.91 & 90.83 & 90.64 & 90.66 & 90.95 & 91.05 & 91.62 \\
\textit{Average \% Entity Coverage by YAGO-4} \\
Train  & 9.78 & 9.63 & 9.46 & 9.97 & 10.01 & 10.03 & 10.01 & 10.03 & 10.01  \\
Validation & 9.80 & 9.62 & 9.51 & 9.98 & 10.03 & 10.10 & 9.68 & 10.02 & 10.15  \\
Test & 9.85 & 9.70 & 9.54 & 10.05 & 10.07 & 10.09 & 10.05 & 10.01 & 10.10 \\
\textit{Avg. \% Known Ents. post Manipulation} \\
Train  & 6.94 & 9.26 & 11.07 & 30.28 & 26.33 & 28.35 & 60.85 & 54.26 & 51.83  \\
Validation & 11.97 & 9.07 & 10.16 & 26.76 & 23.89 & 27.18 & 48.72 & 49.26 & 48.68  \\
Test & 7.68 & 8.99 & 9.03 & 26.13 & 27.15 & 25.76 & 48.85 & 52.72 & 51.51 \\
\bottomrule 
\end{tabular}
\caption{Summary statistics of proposed datasets.}
\label{tab:dataset_stats}
\end{center}
\end{table*}

\subsection{Hyperparameter Search Space for All Detectors}
\label{sec:hyp_space}
Table~\ref{tab:hypspace_compo} displays the search space for hyperparameters used to tune all the detectors.

\begin{table}[htb]
\footnotesize
\begin{center}
\begin{tabular}{l|l} \hline
\textbf{Hyperparameter Name} & \textbf{Hyperparameter Values} \\ \hline 
RoBERTa model variant & Large \\
Minimum frequency of node (i.e., entity) & \{10\} \\
Batch size & \{8\} \\
Initial learning rate & \{1e-5, 2e-5, 3e-5\} \\
Epochs & \{10\} \\
Number of warmup steps & \{10\%\} \\
Node intialization & \{Wikipedia2vec\} \\
Node embedding size &  \{100, 300\} \\
Number of GCN layers & \{1, 2\} \\ \hline
\end{tabular}
\caption{Hyperparameter search space for all detectors.}
\label{tab:hypspace_compo}
\end{center}
\end{table}

\end{document}